# Evaluation of data imputation strategies in complex, deeply-phenotyped data sets: the case of the EU-AIMS Longitudinal European Autism Project

A. Llera[1,2,3], M. Brammer[7], B. Oakley[8], J. Tillmann[6], M. Zabihi[1,2], T. Mei[1,2], T. Charman[5], C. Ecker[5,12], F. Dell'Acqua[7], T. Banaschewski[13], C. Moessnang[11,12], S. Baron-Cohen[14], R. Holt[14], S. Durston[15], D. Murphy[7,8], E. Loth[,7,8], J. K. Buitelaar[1,2,9], D. L. Floris* [1,2,4], C. F. Beckmann* [1,2,16]

1. Donders Institute for Brain, Cognition and Behaviour, Centre for Cognitive Neuroimaging, Nijmegen, The Netherlands.
2. Department of Cognitive Neuroscience, Radboud University Medical Centre, Nijmegen, The Netherlands.
3. LIS data solutions, Machine Learning Group, Santander, Spain.
4. Methods of Plasticity Research, Department of Psychology, University of Zurich, Switzerland.
5. Department of Psychology, Institute of Psychiatry, Psychology, and Neuroscience, King's College London, United Kingdom.
6. Roche Pharma Research and Early Development, Neuroscience and Rare Diseases, Roche Innovation Center Basel, F. Hoffmann–La Roche Ltd., Basel, Switzerland
7. Sackler Institute for Translational Neurodevelopment, Institute of Psychiatry, Psychology, and Neuroscience, King's College London, United Kingdom
8. Department of Forensic and Neurodevelopmental Sciences, Institute of Psychiatry, Psychology, and Neuroscience, King's College London, United Kingdom
9. Karakter Child and Adolescent Psychiatry University Centre, 6525 GC Nijmegen, The Netherlands
10. Department of Psychiatry and Psychotherapy, Central Institute of Mental Health, Medical Faculty Mannheim, University of Heidelberg
11. Department of Applied Psychology, SRH University, Heidelberg, Germany.
12. Department of Child and Adolescent Psychiatry, Psychosomatics and Psychotherapy, University Hospital Frankfurt am Main, Goethe University, Frankfurt, Germany
13. Child and Adolescent Psychiatry, Central Institute of Mental Health, University of Heidelberg, Mannheim, Germany
14. Autism Research Centre, Department of Psychiatry, University of Cambridge, UK
15. Department of Psychiatry, Brain Center Rudolf Magnus, University Medical Center Utrecht, The Netherlands
16. Oxford Centre for Functional Magnetic Resonance Imaging of the Brain (FMRIB), University of Oxford, Oxford, OX3 9DU, U.K.

*Shared last authorship

## Abstract

An increasing number of large-scale multi-modal research initiatives has been conducted in the typically developing population, e.g. [1]–[3], as well as in psychiatric cohorts, e.g. [4]–[7]. Missing data is a common problem in such datasets due to the difficulty of assessing multiple measures on a large number of participants. The consequences of missing data accumulate when researchers aim to explore relationships between multiple measures. Here we aim to evaluate different imputation strategies to fill in missing values in clinical data from a large (total N=764) and deeply characterised (i.e. range of clinical and cognitive instruments administered) sample of N=453 autistic individuals and N=311 control individuals recruited as part of the EU-AIMS Longitudinal European Autism Project (LEAP) consortium. In particular we consider a total of 160 clinical measures divided in 15 overlapping subsets of participants. We use two simple but common univariate strategies, mean and median imputation, as well as a Round Robin regression approach involving four independent multivariate regression models including a linear model, Bayesian Ridge regression [8], as well as several non-linear models, Decision Trees [9], Extra Trees [10] and K-Neighbours regression [11]. We evaluate the models using the traditional mean square error towards removed available data, and consider in addition the KL divergence between the observed and the imputed distributions. We show that all of the multivariate approaches tested provide a substantial improvement compared to typical univariate approaches. Further, our analyses reveal that across all 15 data-subsets tested, an Extra Trees regression approach provided the best global results. This allows the selection of a unique model to impute missing data for the LEAP project and deliver a fixed set of imputed clinical data to be used by researchers working with the LEAP dataset in the future.

**Introduction**

In clinical settings, a broad array of data using questionnaires, observational methods or interviews, and behavioural assessments is acquired that involve a number of individuals ($n$) and a number of clinical variables ($p$). Missing data is a general problem in data analyses [12]–[17] since most algorithms cannot directly handle the presence of missing values. Although there exist probabilistic models able to handle missing observations, these are scarce, strongly tailored for specific analyses and consequently their use is limited and not an standard procedure [18]. Instead, the usual way researchers proceed in such cases is to reduce the sample size ($n$) by removing individuals missing certain data variables resulting in a decrease of statistical power for any further analyses [19]. This problem becomes most notable when performing multivariate analyses involving multiple variables [20], [21], for example classification or clustering, since the number of individuals available in any such analyses will be limited by the simultaneous availability of several clinical measures, reducing the sample size even further. A reduced sample size has a direct effect on the statistical power resulting in reduced sensitivity to and specificity of findings. This is problematic especially in cases where a small effect is usually expected, as it is the case for example in computational psychiatry. At the same time, an increased sample size will also provide more confidence in the observed patterns and increases reproducibility. Other important issues when excluding participants due to one or more missing variables are both the associated 'economic loss' in the sense of not utilising all the (research) resources invested in the study, and the ethical issue of participants' high time investment during data collection, but (partial) data they have provided not being used. Further, data loss can have an even bigger impact on analyses where one wants to study the relationship between different data modalities, such as clinical/behavioural variables and neuroimaging or genetic data [22], [23]. Basically, missing clinical measures reduce the full imaging/genetic sample resulting in a significant loss of statistical power, and a dramatic under-utilisation of investment on the part of funders, researchers and research participants. This is particularly a problem in the case of big-data consortia where a wide range of expensive data collections are performed [1], [6], [7], [24]–[26]. An alternative approach to deal with missing data values is *data imputation* [27]. This approach substitutes missing values by applying a statistical estimation of their values, and consequently avoids reducing the sample size and prevents associated loss issues. A very common and simple strategy for imputation of behavioural or clinical data is substituting individual missing values by the mean or the median of the observed sample values of the respective variable. Even though this approach allows one to retain the original sample size, it does not improve the statistical power of consequent analyses, the reason being that the number of independent clinical observations remains fixed. Furthermore, such simplistic imputation strategies are not well suited when heterogeneity can be expected in the clinical group, e.g. the distribution of observed values is not unimodal. A more advanced strategy which circumvents this shortcoming of mean/median imputations, and is thus able to increase the amount of independent observations, is based on multivariate regression models [28]. These use *all* clinical variables to obtain expectations over the values at each missing value per variable [29]. Such an approach uses a Round-Robin [30], [31] scheduled regression where missing values expectations are iteratively updated through all variables until convergence of all missing values is reached. In such approaches, every missing value expectation for a given variable is different for different participants since it is based on the observations and expectations of all variables for each participant independently. Consequently, this approach increases the number of independent observations with respect to the simpler univariate imputation approaches. Obviously, Round-Robin multivariate regression strategy results are dependent on the regression model chosen, and in fact, this choice is the biggest difference between the most common imputation packages used in practice. For example, some common packages use parametric regression procedures [31], whereas others use non-parametric regression models [32], all cases embedded in a Round-Regression scheduling process.

In this work, we use behavioural/clinical data from the EU-AIMS Longitudinal European Autism Project (LEAP) consortium – the largest, international multi-centre initiative dedicated to identifying biomarkers in Autism Spectrum Disorder (henceforth 'autism'). To study autism at the neurobiological and genetic level, data were collected from a population of individuals with an autism diagnosis as well as from typically developing (TD) individuals between 6-30 years of age. The sample is deeply phenotyped with an extended battery of behavioural, cognitive and clinical assessments alongside a wide range of biomarker measurements such as electroencephalogram, structural and functional magnetic resonance imaging, biochemical markers and genomics [6]. In the LEAP sample in particular, and in most large-scale imaging consortia in general, missing behavioural and clinical data has a large impact due to the extensive and expensive battery of imaging and genetic data acquired. Consequently, clinical data imputation has shown itself necessary to fully exploit the potential of such a rich and costly dataset. The need becomes even more evident in the context of longitudinal study designs such as LEAP, where missing behavioural and clinical data at one timepoint poses additional challenges for meaningful longitudinal analyses. Here, we thus perform a systematic and extensive evaluation of different imputation models to be able to provide a state-of-the-art imputation procedure for the EU-AIMS LEAP cohort in particular and provide a unique set of imputed data to use for all researchers involved in LEAP, thus avoiding biases resulting from different researchers using different models to impute clinical data for their individual analyses. Since the evaluation of such models is not trivial due to the potential non-randomness of missing data, we develop quantitative measures to assess the quality of the imputation.

## Methods

*The dataset*

EU-AIMS LEAP is the to-date largest multi-centre, multi-disciplinary observational study on biomarkers for autism involving a large sample of 764 individuals including 453 autistic children, adolescents and adults and 311 TD individuals (or with mild intellectual disability [ID] without autism) between the ages of 6 and 30 years. Each individual is comprehensively characterised at multiple levels including their clinical profile, cognition, brain structure and function, biochemistry, environmental factors and genomics. This study utilises an 'accelerated longitudinal design', comprising four cohorts defined by age and ability level: Children with either autism or typical development aged 6-11 years and IQ in the typical range, adults with either autism or TD aged 12-17 years and IQ in the typical range, young adults with either autism and TD aged 18-30 years and IQ in the typical range, and adolescents and adults with mild intellectual disability with/without autism aged 12-30 years [6], [33]. The study involves a comprehensive approach to deep phenotyping. Due to differences in age and ability level, measures were divided by experimental design into core measures that were assessed in all participants, and measures that were selectively administered in some schedules which were appropriate for adolescents and/ or adults with higher cognitive function but not for children or those with mild ID. This includes questionnaire measures, such that parents were used as informants in all schedules (except for typically developing adults, where parents were not available to participate in the study) while self-report questionnaires were only used in adolescents and adult. We also aimed to reduce the testing burden of experimental tests (e.g., MRI acquisition times) for children and young people with ID. The full protocol includes a) demographics , such as education of caregiver and parental household income) medical history, b) observational measures of autistic features (e.g., Autism Diagnostic Observation Schedule [ADOS] [34]), c) parent-based interviews (e.g., Autism Diagnostic Interview [ADI-R] [35], adaptive functioning (VABS-II [36]), d) parent- and self-reported questionnaires of the core autism phenotype (e.g., SRS-2 [37]; RBS-R [38]; SSP [39]), associated features (e.g., Sleep Habit Questionnaire [40], Empathy Quotient [41]–[43]; Child Health and Illness Profile[44] and measures of commonly co-occurring conditions (e.g., Attention-Deficit/Hyperactivity Disorder [ADHD]: DSM-5 ADHD rating scale; SDQ [45]; DAWBA [46], anxiety: Beck Anxiety Inventory [47], depression: Beck Depression Inventory [48]. We deliberately included several questionnaires that overlapped in their construct content, e.g. assessing core features of autism, to validate them externally. This means that high correlations between some measures were expected. The protocol further includes e) cognitive assessments, including e.g., Intellectual functioning [IQ]: Wechsler Intelligence Scale for Children [WISC] [49], Wechsler Adult Intelligence Scale [WAIS] [49] handedness: Edinburgh Handedness Inventory [50], social cognition, (e.g., theory of mind: animated shapes task [51]; false belief task [52]) ;executive function Spatial Working Memory [53]. Some cognitive tests used behavioural response variables while others also acquired functional brain responses (e.g., using fMRI Flanker task [54], Social and Non-Social Reward task [55], or EEG (e.g., mismatch negativity, face processing). A detailed description of the clinical cohort and extended characterisation can be found in [6], [33] . In this paper we consider a set of 160 clinical measures in total. A complete list of all included measures in the analyses is provided as Supplementary Table 1 (ST1).

The 160 measures considered in this paper expand self and parent reported measures, and include a subset of measures acquired for all 764 participants, a subset acquired for all 453 individuals with autism, and several other subsets of measures acquired uniquely for subsets of individuals defined by four different enrolment schedules (adults, adolescents, children or intellectual disability [ID]. This resulted in a total of 15 different subsets structured based on group (autism vs. TD), schedule and acquisition method. A summary of all these subsets of participants for which measures are present is summarized in Table 1, where a total of 15 different subsets of individuals and measures are defined. A summary of the number of variables (p), individuals (n), percentage of missing samples as well as the target group (i.e., diagnostic group and enrolment schedule) in which the measure was supposed to be acquired in the first place (i.e., green vs. not acquired in the group=red).

| Subset | Variables (p) | n | n/p | % missing | Groups | Schedules | | | |
|---|---|---|---|---|---|---|---|---|---|
| | | | | | | Adult | Adolescents | Children | ID |
| 1 | 28 | 764 | 30.6 | 16.1 | ASD | | | | |
| | | | | | TD | | | | |
| 2 | 8 | 453 | 56.6 | 12.5 | ASD | | | | |
| | | | | | TD | | | | |
| 3 | 30 | 653 | 21.8 | 27.9 | ASD | | | | |
| | | | | | TD | | | | |
| 4 | 4 | 560 | 140 | 17.9 | ASD | | | | |
| | | | | | TD | | | | |
| 5 | 4 | 653 | 163.3 | 32.8 | ASD | | | | |
| | | | | | TD | | | | |
| 6 | 32 | 478 | 14.9 | 36.7 | ASD | | | | |
| | | | | | TD | | | | |
| 7 | 6 | 458 | 76.3 | 17.7 | ASD | | | | <18 yo |
| | | | | | TD | | | | <18 yo |
| 8 | 1 | 201 | 201 | 31.3 | ASD | | | | >18 yo |
| | | | | | TD | | | | >18 yo |
| 9 | 2 | 235 | 117.5 | 19.2 | ASD | | | | <18 yo |
| | | | | | TD | | | | <18 yo |
| 10 | 14 | 255 | 18.2 | 37.1 | ASD | | | | |
| | | | | | TD | | | | |
| 11 | 6 | 223 | 37.2 | 26.7 | ASD | | | | |
| | | | | | TD | | | | |
| 12 | 14 | 175 | 43.8 | 19.1 | ASD | | | | |
| | | | | | TD | | | | |
| 13 | 8 | 111 | 13.9 | 41 | ASD | | | | |
| | | | | | TD | | | | |
| 14 | 2 | 57 | 28.5 | 44.7 | ASD | | | | >18 yo |
| | | | | | TD | | | | >18 yo |
| 15 | 1 | 334 | 334 | 35.9 | ASD | | | | |
| | | | | | TD | | | | |

Table 1: All clinical data from the EU-AIMS LEAP consortium acquired at wave 1 is summarised as 15 different subsets as indicated in each row. The columns show the number of variables and participants included on each of these subsets as well as the percentage of missing data. Color-coded columns indicated the availability (green) or lack of data (red) as acquired for a subgroup of the participants as indicated in each column. Abbreviations: ASD= autism spectrum disorder, TD=typically developing individuals, ID=intellectual disability.

In Figure 1, we show the correlation structure of all these variables, grouped by subsets as indicated by the horizontal and vertical black lines. We observe that some subsets do not share participants (white areas), and also that many measures are intercorrelated inside and across subsets, providing a primary motivation for multivariate imputation strategies. More detailed information about the variables included in each of these subsets can be found in Supplementary Table 1.

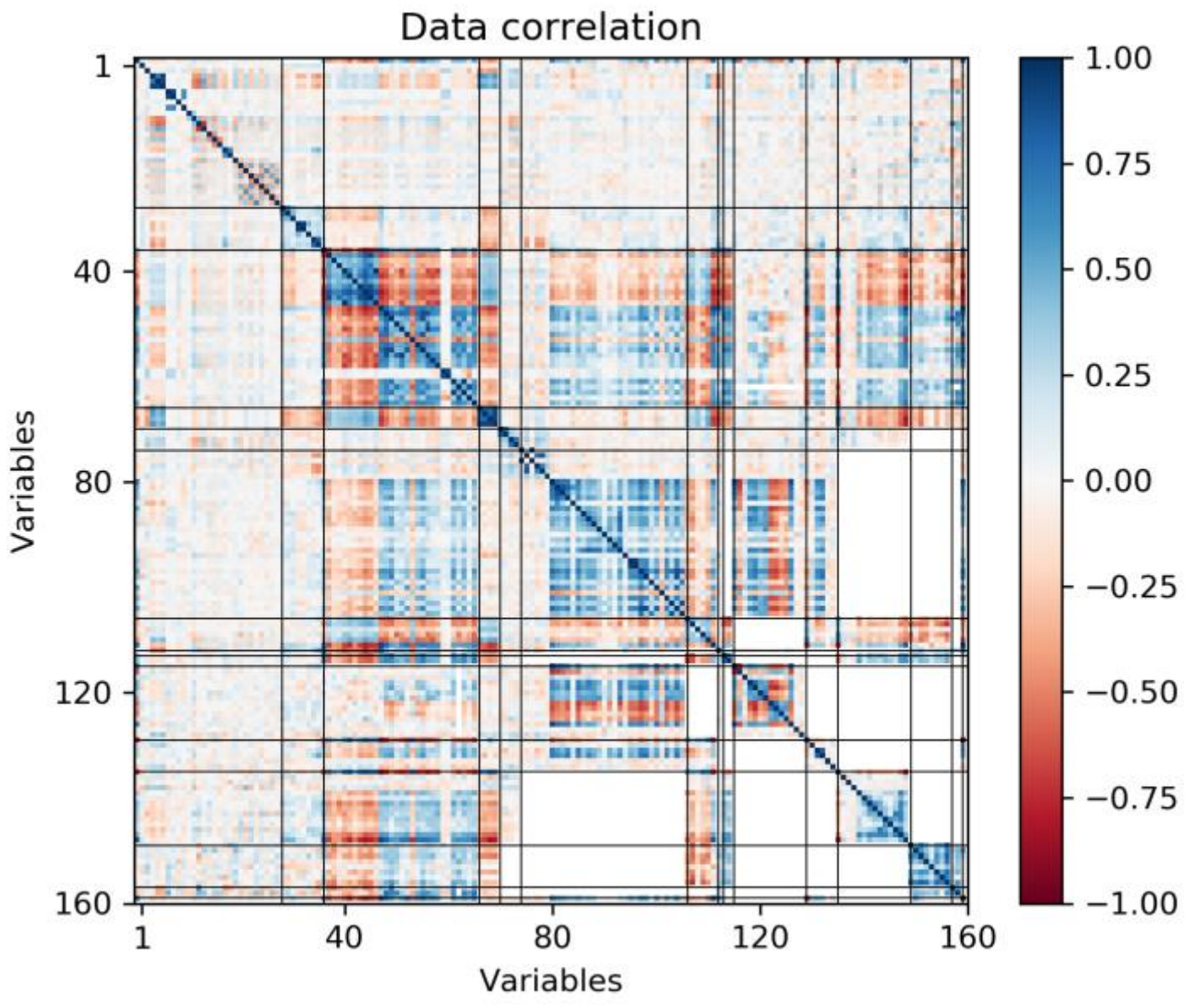

Figure 1: Correlation structure of the 160 clinical measures. White areas correspond to subsets of measures with no shared participants.

There are 28 core clinical measures that comprise all 764 individuals (subset 1), and these measures include for example, age, sex, IQ or handedness. In subset 2, we observe that there are 8 measures comprising all the 453 autistic individuals which include ADOS and ADI. Subset 3 comprises 653 participants and includes all TD individuals along with all autistic children and adolescents; it includes 30 measures with some examples being repetitive behaviour or short sensory profile measures. Subset 4 excludes also autistic adolescents from subset 3 and involves the Vineland Adaptive Functioning Scale. Subset 5 includes TD and autistic individuals, but excludes individuals with ID; this includes a total of 653 individuals and 4 cognitive task measures involving Hariri and theory of mind tasks. Subset 6 excludes all children from subset 5, resulting in a total of 478 individuals and 32 clinical measures as for example Flanker or Social Responsive Scale tests. Subset 7 is also acquired for TD and autistic individuals but excludes adults and individuals with ID older than 18 years, including a total of 458 participants and 6 measures, such as CSBQ and CHIP questionnaires. Without need for further specification of the details for the remaining subsets, it is clear that the individuals included in any of these subsets, are also partially contained in other subsets, and the full picture is a complex organisation of participants and measures (based on diagnostic group, schedule and acquisition type). As a consequence of such a complex structure of clinical data gathering, one cannot use all measures for direct imputation of all the other ones since it would make no sense to impute data that was not supposed to be acquired in a certain group at the first stage which would result in bias. For example, it would not make sense to impute ADI or ADOS measures in TD individuals, as in this study we did not attempt to acquire ADI and ADOS on the TD participants. It is important to note that these 15 subsets of clinical measures have very different properties. First, in terms of the ratio of observations to number of variables, $n/p$ (see table 1). As such, the performance of any regression model can be expected to be different on each subset, even in the hypothetical case of non-missing data. For completeness let's remember that a higher $n/p$ ratio allows more robust and reliable learning [56], [57]. Second, higher percentage of missing values makes the estimation of the missing values harder.

In Figure 2 we visualize some characteristics of the missing data itself, with each row presenting one of the 15 subsets. The left column illustrates the missing values themselves as blue dots, with participants represented in the x-axis and the number of variables included on that subset of the full data in the y-axis. For example, we can observe that subset 1 contains a few measures with no missing values (rows with no blue dot) which include diagnosis, age and sex. In general, for all subsets we can appreciate that white vertical lines show individuals with many variables acquired, while white horizontal lines index measures acquired for many individuals.

In the second and third columns, we color-coded the percentage of shared missing variables between each pair of individuals and the percentage of shared missing individuals between each pair of variables respectively. In these two columns, darker coloured areas index pairs of individuals or measures with many missing shared values respectively. Fourth and fifth columns present histograms of the number of individuals and variables missing respectively. The sixth column presents the correlation between the variables on each subset, where the non-diagonal images show the correlated structure on these measures which motivates the use of multivariate models to estimate their missing values also on each of the subsets independently.

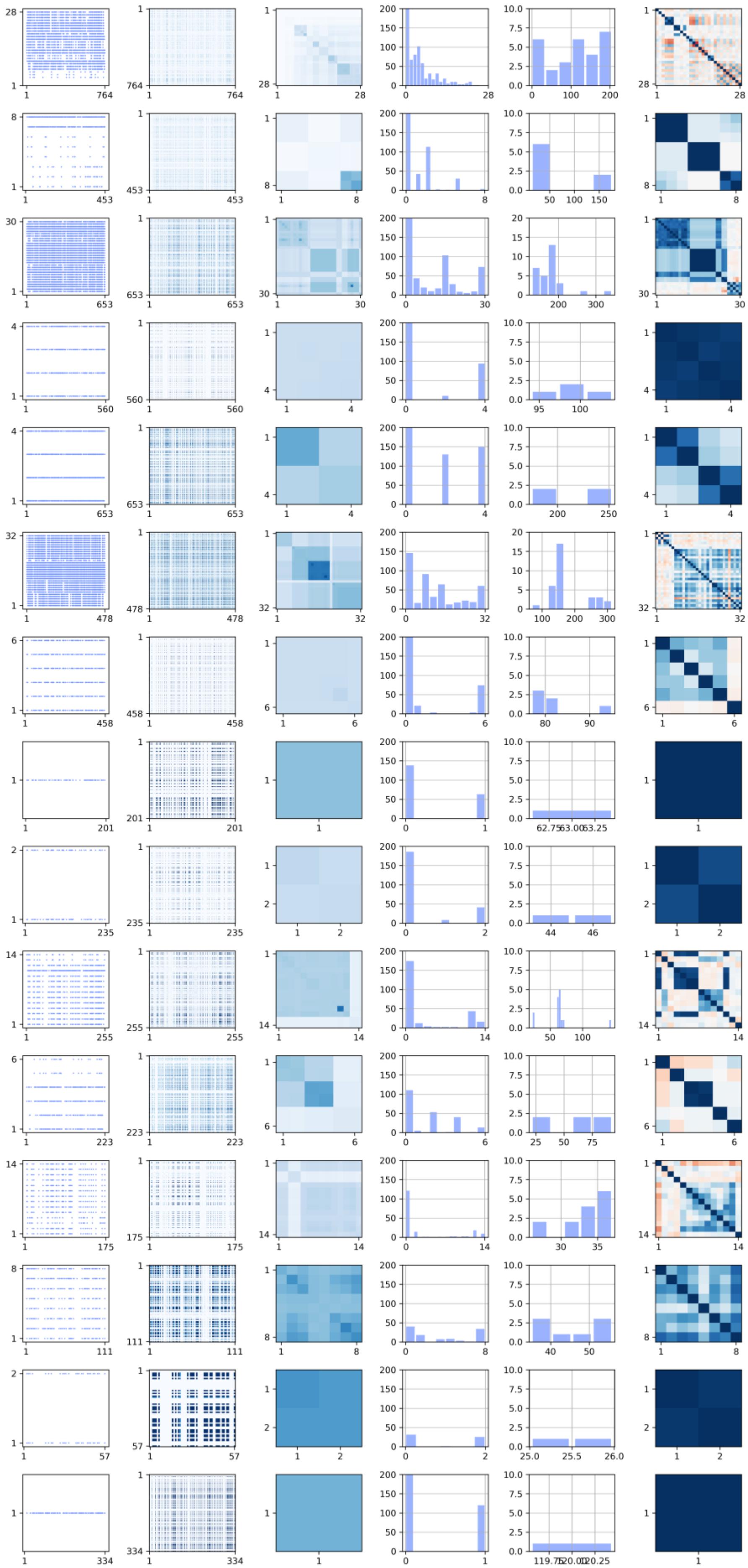

Figure 2: Each row presents information about one of the fifteen subsets. The first column (left) presents missing data as blue dots with individuals presented in the x-axis and number of clinical measures in the y-axis. The second and third columns present the percentage of shared missing variables per pair of individuals, and the percentage of missing individuals per pair of behavioural measures respectively, with darker colours coding an increased percentage.

The fourth and fifth columns present histograms showing the number of individuals missing a number of variables, and the number of variables being missed by a number of individuals. The sixth column present the correlation structure inside each of the subsets i.e. diagonal subsquares of Figure 1.

*Imputation strategies*

For the remainder of this paper we denote by $n$ the number of individuals, by $p$ the number of variables, and by $m$ the number of missing values, where $m = \sum_{j=1}^{p} m_j$ and $m_j$ denotes the number of missing observations for the $j - th$ variable. Consequently, we consider the imputation of a data matrix $D \epsilon M_{nxp}$ where there are $m$ missing values and we denote as $D^*$ the imputed data matrix. We consider the use of six imputation strategies including two simple but common univariate strategies, mean and median imputation, as well as four multivariate regression models including a linear model, Bayesian Ridge (BR) regression, as well as several non-linear models, Decision Trees (DT), Extra Trees (ET) and K-Nearest Neighbors (NN). Table 2 provides an overview of these models.

The univariate imputation strategies substitute all the missing observations at each variable $j \epsilon \{1 \dots p\}$ by some relevant summary statistics at the non-missing values at that variable, i.e. some statistics at the available entries at the $j - th$ column of $D$. In particular here we consider the mean and median imputation strategies.

Such strategies are suboptimal from both a statistical and a clinical point of view; from a statistical point of view they ignore the correlation of the data shown in Figures 1 and 2, and from a clinical point of view, since we know that autism, as many other neurodevelopmental and neuropsychiatric conditions, is clinically and etiologically heterogeneous, meaning that we already a priori assume that there are different relationships between clinical variables and underpinning mechanisms in potentially different subgroups.

These facts strongly motivate moving towards multivariate models for imputation. In the case of multivariate methods, since all variables are needed for imputation of each single variable missing values, we use a Round-Robin [30] regression approach, treating every variable as an output in turn. This approach requires defining an order for variable imputation and for simplicity here we consider an ordering where variables are imputed in an ascending order of number of missing values. Initially, once the first variable of interest to be imputed is selected according to the chosen variable ordering, all other variables missing data values are set to its expectation using mean imputation, and the considered multivariate regression model is used to obtain an expectation of the missing values on the variable of interest. Then the next variable of interest is selected according to the ordering and its originally missing values are estimated as above. The process is repeated for all variables to close the first round of the Round-Robin iterative process and obtain estimations for all missing values that are consequently different from the initial mean imputation values assigned. Then the Round-Robin cycle is repeated as many times as needed, using at each round the estimated missing values from the previous round, till all imputed values at all variables converge. Here we set to 100 the maximum number of Robin-Rounds to perform. All imputations were performed using publicly available tools [58].

Table 2: Imputation strategies considered

| | | **Imputation Strategy** |
|---|---|---|
| Univariate | | Mean |
| | | Median |
| Multivariate Regression with Round-Robin schedule | Linear | Bayesian Ridge (BR) |
| | Non-linear | Decision Trees (DT) |
| | | Extra Trees (ET) |
| | | K-Neighbours (NN) |

*Evaluation*

There is need for a strict validation of the imputation results since the imputation choice can have a strong bias effect on the clinical-brain/genetics associations which need to be minimized. To quantify the quality of each imputation model we use two different measures.

1) We first compute the quality of the imputation using a leave-one-observation-out cross-validation approach. More exactly, for each imputation model, we perform $(nxp) - m$ imputation problems, where at each of the problems we add an extra missing value to the original problem, let's say at location $(i, j)$, resulting in a data matrix to be imputed with m+1 missing values. For clarity of notation lets clarify that for fixed $j \in \{1, \ldots, p\}$, $i \in O_j = \{k_{j,1}, \ldots, k_{j,n-m_j}\}$ with $O_j$ denoting the set of indexes of the available observations for the $j - th$ variable in $D$. Then we perform the imputation using any selected imputation model to obtain an imputed data matrix $D^*$ and compute the total error at the removed value $D_{ij}$ as

$$E(i, j) = sqrt\sqrt{(D_{ij} - D_{ij}^*)^2} \;\; .$$

To have a measure of error considering the scale of each variable independently, we compute a relative error (RE) measure by dividing the total error (E) by the mean of the observed values at $D$ per each variable independently, so

$$RE(i, j) = \frac{E(i, j)}{\frac{1}{n - m_j} \sum_{k \epsilon O_j} D_{kj}}$$

Consequently $RE(i, j)$ is simply a scaled version of $e$ that relates to the size of the error with respect to the size of the variable values, and assigns a value of 0 in the case of no estimation error and a value of 1 when the error $(E)$ is of the size of the mean observed value at that variable. Such representation facilitates the comparison of values on $RE$ across variables taking values at different scales. Finally, to summarize $RE$ per variable we take its mean value across the observations at that variable and we denote it as

$$MRE(j) = \frac{1}{n - m_j} \sum_{k \epsilon O_j} E(j, j), \qquad \forall j \epsilon \{1, \dots, p\} \qquad (1)$$

2) We then perform the imputation of the original data matrix $D$ and compute, at each variable independently, the Kullback-Leibler (KL) divergence [57] between the initially observed distribution and the distribution of estimated values at the missing participants. More precisely,

$$KL(p_j || q_j) = \sum_x p_j(x) \log\left(\frac{p_j(x)}{q_j(x)}\right), \qquad \forall j \epsilon \{1, \dots, p\} \qquad (2)$$

where $p_j(x)$ is the distribution of the observed values at the $j - th$ variable and $q_j(x)$ the distribution of the imputed missing values at that same variable. The KL divergence assigns a value of zero to identical distributions, and increasing values to distributions that deviate from each other.

We would like to remark that $RE$ is an appropriate error function in the case that missing values occur at random, since the cross-validation approach would reproduce perfectly the imputation problem. However, in cases where the missing data does not occur randomly, due to the iterative nature of the multivariate imputation strategies, it might be 'easier' to properly impute the artificially created missing values because one may not rely on expected values for other variables but rather on real observations. Figure 1 columns 2 and 3 suggest that data is not missing at random (it shows rather structured matrices), which motivates the introduction of the second measure of error, the KL divergence, that will penalize models providing distributions at the missing values that deviate from the observed distribution.

It is to note that although each of these measures is informative for each variable, they cannot be studied together since they belong at different scales. However, we can build a proper two-dimensional error function by considering the MRE and KL values per variable relative to some reference model. Consequently, to be able to consider simultaneously the MRE and the KL measures of error, and to be able to pull many variables together to draw any conclussion, we define as a reference model the mean imputation model, and divide for each variable, the MRE and the KL measures at each model by the MRE and KL values obtained by the mean imputation model. In this way, we obtain an *MRE and KL measures relative to the mean imputation*, assigning for each variable the mean imputation performance to the plane point (1,1), and all other performances can be pulled together as they represent a relative improvement with respect to the mean imputation. Consequently, for a given variable and a fixed imputation model, we consider the frobenious norm of such two-dimensional 'error vector' as a global measure of error that combines both MRE and KL.

*Order of imputation*

As we showed in Table 1, the considered clinical data breaks down to a very complex organization of measures according to the population for which they are acquired, that can be summarized as 15 different subsets of data. Consequently, we cannot use all measures for imputation of all the other measures since it would not be sensible to impute for certain individuals measures that were not intended to be acquired for them in the experiment design. However, imputation of each of the 15 subsets independently would be suboptimal since we observed correlations also across subsets in Figure 1. Consequently, one needs to combine subsets to maximise the imputation power. To that end, we performed an exhaustive search to find the optimal order of imputation of each of these subsets, while for imputation of a target subset we used any previously imputed subsets, as long as the target population is contained in the previously imputed subsets.

| order | input | output | Conditioned to |
|---|---|---|---|
| 1st | Subset 1 | Subset 1* | None |
| 2nd | Subset 3 | Subset 3* | Subset 1* |
| 3rd | Subset 4 | Subset 4* | Subsets 1*, 3* |
| 4th | Subset 2 | Subset 2* | Subsets 1*,3*,4* |
| 5th | Subset 5 | Subset 5* | Subset 1* |
| 6th | Subset 6 | Subset 6* | Subset 1*,5* |
| 7th | Subset 7 | Subset 7* | Subsets 1*, 3* |
| 8th | Subset 15 | Subset 15* | Subsets 1*,3* |
| 9th | Subset 8 | Subset 8* | Subsets 1*,3*,4* |
| 10th | Subset 9 | Subset 9* | Subsets 1*,3*,4*,7* |
| 11th | Subset 10 | Subset 10* | Subsets 1*,5*,6* |
| 12th | Subset 11 | Subset 11* | Subsets1*,3*,5*,6*,7*,15* |
| 13th | Subset 12 | Subset 12* | Subsets 1*,3*,4*,5*,7*,9* |
| 14th | Subset 13 | Subset 13* | Subsets1*,3*,4*,15* |
| 15th | Subset 14 | Subset 14* | Subsets1*,3*,4*,8*,13*,15* |

Table 3: Order followed for imputation of the subsets. The last column shows the imputed subsets used for imputation of each subset indicated in the second column.

The process starts with the imputation of subset 1 in isolation, since all participants were planned to be measured with respect to these 28 variables. It is important to mention that from subset 1 we removed the clinical measure diagnosis to not bias the imputation towards the diagnosis label and to not produce a bias effect in any posterior study on these imputed data. Our brute force optimization showed that the next subset to impute it should be subset number 3, which is acquired for all participants with the only exception of autistic adults; for imputation of subset 3 we used the imputed values of subset 1, restricted to the individuals in subset 3, in addition to the variables on subset 3. After we proceeded to subset 4 and then to subset 2. In Table 3 we provide the structure of the ordering performed to maximize the power of all the imputation process, where an asterisk denotes an imputed file. The fourth column indicates the already imputed files that are considered for imputation of each input file.

**Results**

Following the ordering of the 15 subsets of clinical measures indicated in Table 3, we proceeded to the imputation of the missing values in the clinical dataset from EU-AIMS LEAP. As illustrated in section 'Methods: The dataset', each of these data matrices present different challenges to perform their imputation, with for example subset 6 being more challenging than subset 2, since the subset has a smaller $n/p$ ratio and has many more missing values (see Table 1). Consequently, these 15 subsets serve as an interesting test bed to study the robustness of the different algorithms in general and not uniquely for this dataset, since we can check the performance in the harder problems in relation to the simpler ones.

Figure 3 shows the MRE and KL plane relative to the mean imputation for each subset (subplots), with each dot representing one clinical variable in that subset, the different imputation models being color-coded and the colored squares representing the mean of the values for a given model in that subset. Further, the bottom right figure shows the mean performance of each model pulled across all measures of all 15 subsets. Recap for interpretation that models that are lower with respect to the y-axis perform better with respect to the KL divergence, while models that are plotted more to the left with respect to the x-axis perform better with respect to the MRE measure. Globally, models closer to (0,0) perform better. We first observe that in general the mean and median imputation perform much worse than all other models with respect to the MRE and also to the KL divergences i.e. blue and yellow dots are further from 0 with respect to both axis. This is clear evidence for superior performance of multivariate models for such clinical measures imputation. With respect to the multivariate models we appreciate that NN performs well with respect to the KL, which makes sense since by looking at some of the closest neighbours its allowed to sample the full space and get a distribuion closer to the initially observed one. However, KNN fails to provide a robust improvement with respect to the RME, and in some subsets is even worse than the mean imputation (red squares not appearing in figure, for example for subset 13). From the remaining three models, we observe that Extra Trees Regressor (purple) and Bayesian Ridge Regression (green) outperform Decission Trees (brown). Although both Extra Trees and Bayesian Ridge provide an impresive improvement with respect to the mean imputation in terms of RME (~40 % reduction of error), Extra Trees provides a bigger improvement with respect to the KL divergence (~75 vs ~55 % reduction of KL). Another interesting observation is that the imputation of all subsets provide a similar pattern of organization of the models performances, showing the robustness of the model performances across all subsets. This is a interesting finding given the huge differences in the $n/p$ ratios as well as in the number of missing obervations on each subset (Table 1). This representation confirms that the median imputation provides a similar performance to the mean imputation and they are the worst models. It further shows that that BR provides in general a very high relative MRE improvement, but a lower relative KL improvement than the other multivariate models. It further highlights that the Extra Tree regressor is the model performing best in expectation. In fact, to compare the best two models, a paired t-test between the norms of the 2-dimensional errors in relative KLvsMRE plane of the ET and the BR models showed a signifcantly reduced error in favor of the ET model ($t=4{,}01$, $p<9 \times 10^{-5}$).

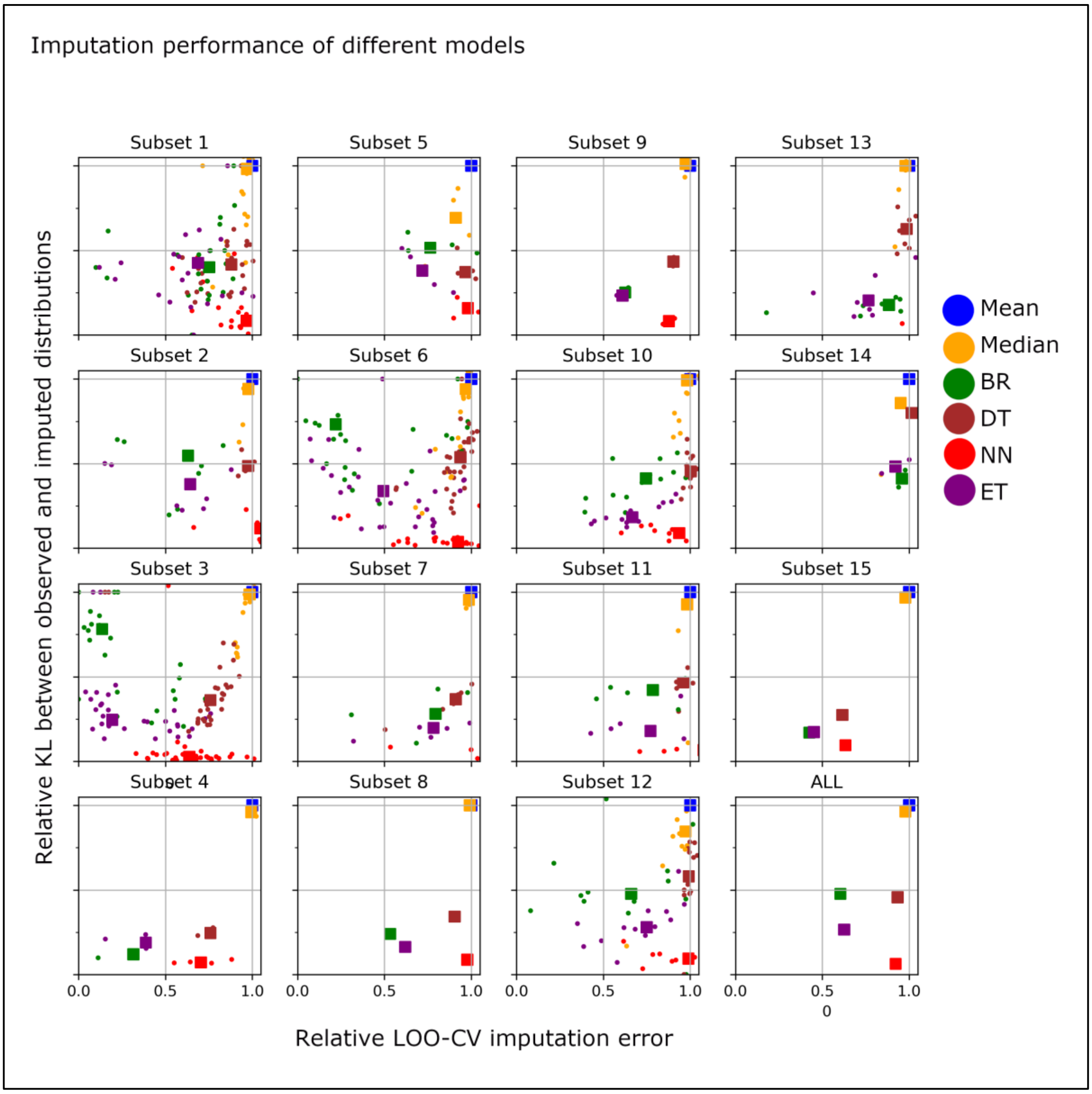

Figure 3: Visualization of the imputation performance at the clinical measures acquired at each of the subsets. Each subfigure presents the performance for each clinical measure in the subset as dots, and for the 6 imputation models considered (color coded). The colored squares show the mean across measures per model. For each subset, the x-axis shows the mean imputation error (MRE) relative to the mean imputation model, and the y-axis the KL-divergence between the distribution at the availabel (observed) data and the imputed data at the missing values, again relative to the mean imputation model. Color coding in the legend: blue and yellow represent the univariate models, mean and median imputation respectively; green represents a multivariate linear Bayesian Ridge regression model (BR). The remaining colors encode multivariate non-linear models, with brown encoding decission trees (DT), red encoding k-nearest neigbours (NN), brown and purple extra tree regressors (ET).

**Discussion**

We performed a comprehensive analysis and evaluation of six different imputation methods to compare the weaknesses and strengths of different methodologies to perform imputation of clinical variables. To that end we used 15 different subsets of clinical variables from the EU-AIMS LEAP dataset that have considerable differences in terms of ratio between number of variables and number of observations (n/p) as well as in terms of percentage of missing data values. We used standard univariate imputation techniques, i.e. mean and median imputation, as well as several multivariate regression models, i.e. Bayesian Ridge, Random Forest, Extra Trees, Decision trees. All the multivariate models were involved in a Round-Robin iterative scheduling till convergence of all missing values estimations. We evaluated the imputation using two different error measures, computing the error at the originally observed data using a leave-one-observation-out cross-validation approach, and also by computing the KL-divergence between the observation distributions and the imputed value distributions at each variable independently. To be able to compare the results of all models we scaled both error measures with respect to the mean imputation performances to obtain a measure of improvement with respect to the simplest mean imputation model. Even though the considered subsets had very different characteristics, the expected improvement with respect to the simpler mean imputation resembled in both cases a very similar pattern showing that the models performed in a similar fashion at the simplest as well as the hardest/most complex scenarios. In particular we observed that Extra Tree Regression was in expectation the best model for imputation of this dataset. All models were initially independently evaluated using grid search in a set of model parameters and the solution with the best set of parameters per model was selected and presented in this paper. In particular, for the Extra Tree Regression model we found that a model with 10 trees provided the best solution. It is to note that the Round-Robin regression approach is also implemented in the R-package for imputation MICE [31] and in fact, the python package we used here for imputation [58] is inspired in MICE. A particularity of MICE is that it models categorical variables using logistic or multinomial regression and continuous variables using linear regression [59]. As such MICE has more flexibility than the presented Bayesian Ridge Regression model, since is tailored to model specifically categorical variables. However, the Tree based methods we considered are also able to automatically capture such categorical structure from the data, and also non-linearities between all the variables or handle multimodal discrete and also continuous distributions that might directly scape to MICE, or require strong modelling and data domain specific knowledge. This has been empirically shown in [60] where it was found that although the difference between tree based methods and parametric MICE is not big, tree based methods outperformed the parametric models. Note that to handle multimodal distributions is necessary for variables where high heterogeneity is observed and of outmost importance in the autism research where stratification based on clinical and imaging data is expected. One added particularity of MICE is that it runs the imputation problem many times with different initializations, returning finally the average of these imputations as final value. The most interesting of this approach is that it provides the standard deviation over the imputed values which serves as a measure of reliability in the imputation. However, one might argue that some random initializations might also deviate from a reasonable estimation, resulting in a biased expectation value. Note that our extensive analyses also perform a validation that allows to get a measure of the quality of the imputation at each variable as given by the MRE and the KL divergences. However, we also considered a multiple imputation scenario for the best of our models, the Extra Tree Regressor. As suggested [58] we did not change the mean imputation as initialization but we rather used 100 different seeds to initially randomly build the regression trees. The results showed a standard deviation of order $10^{-3}$ at all the variables, showing that the estimation obtained using Extra Tree Regressors is extremely robust. Another similarity between the models employed in this work and well known models commonly used come from Random Forest regression embedded on Round-Robin scheduling being equivalent to another common package, missForest [32]. Although we did not include the full evaluation of Random Forest in this work, we performed several analyses during the preliminary preparation of this work and we observed that it would not improve ET or BR, its convergence was less satisfying, and the computational cost was orders of magnitude bigger. Our choice of software is driven by the flexibility of the packages to implement several regression models within the same framework, making the comparison between different models simpler and less error prone. We believe that the choice of model, and not of software, is critical for the quality of the imputation.

The Round-Robin scheduling procedure requires defining a variable ordering for imputation, and although here we report results using an increasing number of missing observations for variable ordering, results using a decreasing order did show similar results, both in terms of squared error and in terms of KL divergences between the observed and the imputed distributions at most variables, and for most models. Also, the patterns of model performances were identical. In conclusion, we systematically searched the best practice scenario for imputation of the clinical variables in this sample and found that Extra Trees Regressor (🌳🌲🌴) was in expectation the best model. Given the different characteristics of the 15 data samples we consider that these results might also extrapolate to different datasets. As a result of this analyses we deliver the tools for imputation comparison we developed at https://github.com/allera/Imputation, and deliver imputed data to the EU-AIMS LEAP consortium.

An arising question is whether we can synthetically generate other missing measurements from such big data consortiums as for example structural brain images. The presented models are useful in their own for different types of vector data, however, models implementing spatial constrains should be more appropriate to interpolate data where a clear non-isotropic spatially smooth 3d distribution is expected. Ongoing research focuses on the imputation of missing T1w images, using existing T1w images and behavioural readouts, e.g. age, sex, weight. To that end we are considering extended CNNs and we expect to be able to, for example, generate synthetic T1w images with smaller brain volume for younger participants. Once more, the quality of this approach can be validated by removing participants one at a time and checking the quality of the recovered image. Even more, given the relationship between structural features and functional features extracted from fMRI [22], we also aim to predict expected functional features based on structural and behavioural readouts, also using spatial convolution models. Such results are expected to follow up this work.

**Acknowledgment**

We thank all participants and their families for participating in this study. We also gratefully acknowledge the contributions of all members of the EU-AIMS LEAP group: Jumana Ahmad, Sara Ambrosino, Bonnie Auyeung, , Sarah Baumeister, Sven Bölte, Thomas Bourgeron, Carsten Bours, Daniel Brandeis, Claudia Brogna, Yvette de Bruijn, Bhismadev Chakrabarti, Ineke Cornelissen, Daisy Crawley, Guillaume Dumas, Jessica Faulkner, Vincent Frouin, Pilar Garcés, David Goyard, Lindsay Ham, Hannah Hayward, Joerg Hipp, Mark H. Johnson, Emily J.H. Jones, Prantik Kundu, Meng-Chuan Lai, Xavier Liogier D' ardhuy, Michael V. Lombardo, David J. Lythgoe, René Mandl, Andre Marquand, Luke Mason, Maarten Mennes, Andreas Meyer-Lindenberg, Nico Mueller, Laurence O' Dwyer, Marianne Oldehinkel, Bob Oranje, Gahan Pandina, Antonio M. Persico, Barbara Ruggeri, Amber Ruigrok, Jessica Sabet, Roberto Sacco, Antonia San José Cáceres, Emily Simonoff, Will Spooren, Roberto Toro, Heike Tost, Jack Waldman, Steve C.R. Williams, Caroline Wooldridge, and Marcel P. Zwiers. This project has received funding from the Innovative Medicines Initiative 2 Joint Undertaking under grant agreement No 115300 (for EU-AIMS) and No 777394 (for AIMS-2-TRIALS). This Joint Undertaking receives support from the European Union's Horizon 2020 research and innovation programme and EFPIA and AUTISM SPEAKS, Autistica, SFARI. This work has also been supported by the Horizon2020 programme CANDY Grant No. 847818). DLF is supported by funding from the European Union's Horizon 2020 research and innovation programme under the Marie Skłodowska-Curie grant agreement No 101025785. This work was also supported by the Netherlands Organization for Scientific Research through Vidi grants (Grant No. 864.12.003 [to CFB]). The research leading to the presented work has received funding from the developing Human Connectome Project (dHCP) through a Synergy Grant by the European Research Council under the European Union's Seventh Framework Programme (FP/2007-2013), ERC Grant Agreement no. 319456. We further gratefully acknowledge support from the Netherlands Organization for Scientific Research (NWO) through VIDI grant to CFB (864.12.003). We also gratefully acknowledge funding from the Wellcome Trust UK Strategic Award (098369/Z/12/Z).

The funders had no role in the design of the study; in the collection, analyses, or interpretation of data; in the writing of the manuscript, or in the decision to publish the results. Any views expressed are those of the author(s) and not necessarily those of the funders.

**Disclosures**

JKB has been a consultant to, advisory board member of, and a speaker for Takeda/Shire, Medice, Roche, and Servier. He is not an employee of any of these companies and not a stock shareholder of any of these companies. He has no other financial or material support, including expert testimony, patents, or royalties. CFB is director and shareholder in SBGneuro Ltd. TC has received consultancy from Roche and Servier and received book royalties from Guildford Press and Sage. DM has been a consultant to, and advisory board member, for Roche and Servier. He is not an employee of any of these companies, and not a stock shareholder of any of these companies. TB served in an advisory or consultancy role for Lundbeck, Medice, Neurim Pharmaceuticals, Oberberg GmbH, Shire, and Infectopharm. He received conference support or speaker's fee by Lilly, Medice, and Shire. He received royalities from Hogrefe, Kohlhammer, CIP Medien, Oxford University Press; the present work is unrelated to these relationships. JT is a current full-time employee of F. Hoffmann–La Roche Ltd. The other authors report no biomedical financial interests or potential conflicts of interest.

**Supplementary table 1**

| Sub-set | n | p | Clinical/cognitive/demographic measure | Sub-scale | Acquisition | Observations | missing (%) |
|---|---|---|---|---|---|---|---|
| Subset 1 | 764 | 1 | diagnosis | / | clinical assessment | 764 | 0 |
| | | 1 | sex | / | self/parent | 764 | 0 |
| | | 1 | age | / | self/parent | 764 | 0 |
| | | 3 | WASI | Full scale IQ | clinical assessment | 751 | 1,7 |
| | | | | Verbal IQ | | 747 | 2,23 |
| | | | | Performance IQ | | 753 | 1,44 |
| | | 1 | Edinburgh Handedness Inventory | Handedness | self/parent | 621 | 18,72 |
| | | 2 | Months of education | mother | self/parent | 641 | 16,1 |
| | | | | father | | 644 | 15,7 |
| | | 1 | Annual household income | income | self/parent | 559 | 26,83 |
| | | 2 | Block design | Differnce score RT | cognitive assessment | 598 | 21,73 |
| | | | | Differnce score accuracy | cognitive assessment | 598 | 21,73 |
| | | 2 | Spatial working memry | Total between search error | cognitive assessment | 706 | 7,59 |
| | | | | Total within search error | | 706 | 7,59 |
| | | 1 | Theory of Mind - Animated Shapes task | Accuracy | cognitive assessment | 634 | 17,02 |
| | | 2 | Theory of Mind - False Belief Task | False belief continuous score | cognitive assessment | 634 | 17,02 |
| | | | | Egocentric bias score | | 634 | 17,02 |
| | | 3 | Probabilistic Reversal Learning | Proportion persevarative error | cognitive assessment | 664 | 13,09 |
| | | | | Lose-shift | | 664 | 13,09 |
| | | | | Win-stay | | 664 | 13,09 |
| | | 4 | Non-social reward | win reaction time | fMRI task | 569 | 25,52 |
| | | | | win accuracy | | 573 | 25 |
| | | | | neutral reaction time | | 559 | 26,83 |
| | | | | neutral accuracy | | 573 | 25 |
| | | 4 | Social reward | win reaction time | fMRI task | 634 | 17,02 |
| | | | | win accuracy | | 640 | 16,23 |
| | | | | neutral reaction time | | 626 | 18,06 |
| | | | | neutral accuracy | | 640 | 16,23 |
| Subset 2 | 453 | 3 | ADOS2 | Calibrated Severity Scores (CSS) | clinical assessment | 440 | 2,87 |
| | | | | Social Affect CSS | | 440 | 2,87 |

| | | | | | | |
|---|---|---|---|---|---|---|
| | | | | Restricted and Repetitive Behaviours CSS | | 440 | 2,87 |
| | | 5 | ADI-R | Social domain | parent | 427 | 5,74 |
| | | | | Communication domain | | 427 | 5,74 |
| | | | | Restricted and Repetitive Behaviours domain | | 427 | 5,74 |
| | | | | Age first single words | | 295 | 34,88 |
| | | | | Age first phrases | | 276 | 39,07 |
| Subset 3 | 653 | 1 | Repetitive Behavior Scale-Revised | Total | parent | 521 | 20,21 |
| | | 10 | Short Sensory Profile (SSP) | Total | parent | 391 | 40,12 |
| | | | | Tactile sensitivity | | 466 | 28,64 |
| | | | | Taste/smell sensitivity | | 444 | 32,01 |
| | | | | Movement sensitivity | | 459 | 29,71 |
| | | | | Underresponsive | | 479 | 26,65 |
| | | | | Auditory filtering | | 510 | 21,9 |
| | | | | Low energy | | 484 | 25,88 |
| | | | | Visual/ auditry sensitivity | | 496 | 24,04 |
| | | | | Hypersensitivity | | 481 | 26,34 |
| | | | | Hyposensitivity | | 491 | 24,88 |
| | | 2 | ADHD rating scale | Inattentiveness | parent | 523 | 19,91 |
| | | | | Hyperactivity/impulsivity | | 523 | 19,91 |
| | | 1 | Social Responsiveness Scale-2 | Raw Score | parent | 523 | 19,91 |
| | | 9 | Strengths & Difficulties Questionnaire (SDQ) | Total difficulties | parent | 460 | 29,56 |
| | | | | Emotional problems | | 460 | 29,56 |
| | | | | Conduct problems | | 460 | 29,56 |
| | | | | Peer problems | | 460 | 29,56 |
| | | | | Prosocial problems | | 460 | 29,56 |
| | | | | Impact problems | | 460 | 29,56 |
| | | | | Internalising problems | | 460 | 29,56 |
| | | | | Externalising problems | | 460 | 29,56 |
| | | | | Hyperactivity | | 460 | 29,56 |
| | | 1 | Columbia Impairment Scale | Total | parent | 495 | 24,2 |
| | | 6 | DAWBA | Externalising Behaviour band | parent | 462 | 29,25 |
| | | | | Internalising Behaviour band | | 462 | 29,25 |
| | | | | Anxiety band | | 443 | 32,16 |
| | | | | Depression band | | 512 | 21,59 |
| | | | | Behaviour Dis band | | 311 | 52,37 |
| | | | | ADHD band | | 512 | 21,59 |
| Subset 4 | 560 | 4 | Vineland-II | Communication Domain Score | parent | 466 | 16,79 |
| | | | | Living Domain Score | | 462 | 17,5 |
| | | | | Socialisation Domain Score | | 456 | 18,57 |
| | | | | Adaptive Behaviour Composite (ABC) standard score | | 456 | 18,57 |
| Subset 5 | 653 | 2 | Hariri | accuracy faces | fMRI task | 399 | 38,9 |
| | | | | accuracy forms | | 399 | 38,9 |
| | | 2 | Theory of Mind | accuracy all | fMRI task | 479 | 26,65 |
| | | | | accuracy | | 479 | 26,65 |
| Subset 6 | 478 | 6 | Flanker | congruent reaction time | fMRI task | 353 | 26,15 |
| | | | | congruent accuracy | | 353 | 26,15 |
| | | | | incongruent reaction time | | 353 | 26,15 |
| | | | | incongruent accuracy | | 353 | 26,15 |
| | | | | neutral accuracy | | 353 | 26,15 |
| | | | | nogo accuracy | | 353 | 26,15 |
| | | 1 | Social Responsiveness Scale-2 | SRS-self | self | 320 | 33,05 |
| | | 4 | Toronto Alexithymia Scale | Describe | self | 331 | 30,75 |
| | | | | Identify | | 331 | 30,75 |
| | | | | External | | 331 | 30,75 |
| | | | | Total | | 331 | 30,75 |
| | | 1 | Columbia Impairment Scale | Total | self | 326 | 31,8 |
| | | 8 | Adult Sleep Habit Questionnaire | bedtime | self | 212 | 55,65 |
| | | | | sleep behaviour | | 188 | 60,67 |
| | | | | waking | | 216 | 54,81 |
| | | | | morning waking | | 218 | 54,39 |
| | | | | sleep habits | | 219 | 54,18 |
| | | | | daytime sleep | | 211 | 55,86 |
| | | | | frequency | | 165 | 65,48 |
| | | | | total | | 209 | 56,28 |
| | | 3 | DAWBA | Internalising Behaviour band | self | 330 | 30,96 |
| | | | | Anxiety band | | 314 | 34,31 |
| | | | | Depression band | | 408 | 14,64 |
| | | 9 | Strengths & Difficulties Questionnaire (SDQ) | Total difficulties | self | 323 | 32,43 |
| | | | | Emotional problems | | 323 | 32,43 |
| | | | | Conduct problems | | 323 | 32,43 |
| | | | | Peer problems | | 323 | 32,43 |
| | | | | Prosocial problems | | 323 | 32,43 |
| | | | | Impact problems | | 323 | 32,43 |
| | | | | Internalising problems | | 323 | 32,43 |
| | | | | Externalising problems | | 323 | 32,43 |
| | | | | Hyperactivity | | 323 | 32,43 |
| Subset 7 | 458 | 1 | CSBQ | CSBQ | parent | 378 | 17,47 |
| | | 5 | CHIP | Mean Satisfaction Score | parent | 381 | 16,81 |
| | | | | Mean Comfort Score | | 381 | 16,81 |
| | | | | Mean Resilience Score | | 381 | 16,81 |
| | | | | Mean Risk avoidance Score | | 378 | 17,47 |
| | | | | Mean Achievement Score | | 363 | 20,74 |
| Subset 8 | 201 | 1 | ASBQ | | parent | 138 | 31,34 |
| Subset 9 | 235 | 2 | Beck Inventory Youth | Depression Inventory | parent | 192 | 18,3 |
| | | | | Anxiety Inventory | | 188 | 20 |
| Subset 10 | 255 | 1 | AQ | adult | self | 190 | 25,49 |
| | | 1 | EQ | adult | self | 187 | 26,67 |
| | | 1 | SQ | adult | self | 189 | 25,88 |
| | | 4 | WHOQOL-BREF | Physical health (raw score) | self | 193 | 24,31 |
| | | | | Psychological (raw score) | | 193 | 24,31 |
| | | | | Social relationships (raw score) | | 192 | 24,71 |
| | | | | Environment (rawscore) | | 111 | 56,47 |
| | | 2 | ADHD rating scale | Inattentiveness | self | 190 | 190 |
| | | | | Hyperactivity/impulsivity | | 190 | 25,49 |
| | | 1 | ASBQ | adult | self | 185 | 27,45 |
| | | 2 | Beck Inventory | Anxiety Inventory | self | 192 | 24,71 |
| | | | | Depression Inventory | | 190 | 25,49 |
| | | 2 | Read the Mind in the Eyes Test- ADULT | mean reaction time correct | cognitive assessment | 232 | 9,02 |
| | | | | percent correct | | 232 | 9,02 |
| Subset 11 | 223 | 1 | EQ | adoescent | parent | 157 | 29,6 |
| | | 1 | SQ | adoescent | parent | 158 | 29,15 |
| | | 2 | Beck Inventory | Anxiety Inventory | self | 130 | 41,7 |
| | | | | Depression Inventory | | 132 | 40,81 |
| | | 2 | Read the Mind in the Eyes Test- Adolescent | mean reaction time correct | cognitive assessment | 202 | 9,42 |
| | | | | percent correct | | 202 | 9,42 |
| Subset 12 | 175 | 2 | Child EQ-SQ | EQ | parent | 142 | 18,86 |
| | | | | SQ | | 142 | 18,86 |
| | | 2 | Read the Mind in the Eyes Test- Child | mean reaction time correct | cognitive assessment | 149 | 14,86 |
| | | | | percent correct | | 149 | 14,86 |
| | | 9 | Child Sleep Habit Questionnaire | daytime sleep | parent | 142 | 18,86 |
| | | | | disordered breathing | | 138 | 21,14 |
| | | | | parasomnias | | 138 | 21,14 |
| | | | | wakings | | 139 | 20,57 |
| | | | | anxiety | | 138 | 21,14 |
| | | | | duration | | 141 | 19,43 |
| | | | | onset delay | | 144 | 18,71 |
| | | | | bedtime resist | | 140 | 20 |
| | | | | total | | 138 | 21,14 |
| | | 1 | Child-AQ | AQ | parent | 143 | 18,29 |
| Subset 13* | 111 | 8 | Adult Sleep Habit Questionnaire* | bedtime | parent | 71 | 36,04 |
| | | | | sleep behaviour | | 59 | 46,85 |
| | | | | waking | | 64 | 42,34 |
| | | | | morning waking | | 75 | 32,43 |
| | | | | sleep habits | | 76 | 31,53 |
| | | | | daytime sleep | | 66 | 40,54 |
| | | | | frequency | | 55 | 50,45 |
| | | | | total | | 58 | 47,75 |
| Subset 14 | 57 | 2 | Beck Inventory Adult D | Anxiety Inventory | parent | 31 | 45,61 |
| | | | | Depression Inventory | | 32 | 43,86 |
| Subset 15 | 334 | 1 | AQ | adoescent | parent | 214 | 35,9 |

*This questionnaire was designed for schedules A (autism), B (autism) and D. We did not include schedule A and B in the imputation here,

as consistently across the total score and the sub-measure more than 85% of data was missing